\documentclass[runningheads]{llncs}
\usepackage[T1]{fontenc}
\usepackage{graphicx}
\usepackage{comment}
\usepackage{amsmath}
\usepackage{amsfonts}
\usepackage{amssymb}
\usepackage{graphicx}
\usepackage{textcomp}
\usepackage{xcolor}
\usepackage{multirow}
\usepackage{subcaption}
\usepackage{algorithm}
\usepackage{algorithmic}
\usepackage{flushend}
\usepackage[misc]{ifsym}

\begin{document}
\title{GDM: Dual Mixup for Graph Classification\\ with Limited Supervision}
%

\author{Abdullah Alchihabi\inst{1} 
\and
Yuhong Guo\inst{1,2} 
}
\authorrunning{A. Alchihabi \and Y. Guo}
%
\institute{School of Computer Science, Carleton University, Ottawa, Canada \and
Canada CIFAR AI Chair, Amii, Canada\\
\email{abdullahalchihabi@cmail.carleton.ca, yuhong.guo@carleton.ca}
}

\toctitle{GDM: Dual Mixup for Graph Classification with Limited Supervision}
\tocauthor{Abdullah~Alchihabi, Yuhong~Guo}
\maketitle              
\begin{abstract}
Graph Neural Networks (GNNs) require a large number of labeled graph samples to obtain good performance on the graph classification task. The performance of GNNs degrades significantly as the number of labeled graph samples decreases. 
To reduce the annotation cost, it is therefore important to develop graph augmentation methods that can generate new graph instances 
to increase the size and diversity of the limited set of available labeled graph samples. 
In this work, we propose a novel mixup-based graph augmentation method, Graph Dual Mixup (GDM), 
that leverages both functional and structural information of the graph instances to generate new labeled 
graph samples. 
GDM employs a graph structural auto-encoder to learn structural embeddings of the graph samples, 
and then applies mixup to the structural information of the graphs in the learned structural embedding space 
 and generates new graph structures from the mixup structural embeddings.  
As for the functional information, GDM applies mixup directly to the input node features of the graph samples 
to generate functional node feature information for new mixup graph instances. 
Jointly, the generated input node features and graph structures yield new graph samples which can supplement the set of original labeled graphs. 
Furthermore, we propose two novel Balanced Graph Sampling methods to enhance 
the balanced difficulty and diversity for the generated graph samples. 
Experimental results on the benchmark datasets demonstrate that our proposed method substantially outperforms the state-of-the-art graph augmentation methods when the labeled graphs are scarce.

\end{abstract}

\section{Introduction}

Graph Neural Networks (GNNs) have successfully tackled a wide range of graph related tasks such as node classification, knowledge graph completion, and graph classification. 
In particular, the graph classification task has been addressed 
using various GNN models such as
Graph Convolution Networks (GCNs) \cite{kipf2017semisupervised}, 
Graph Attention Networks (GATs) \cite{velickovic2018graph}, and Graph Isomorphism Networks (GINs) \cite{xu2018how}. 
The effectiveness of such GNN models
can be attributed to their
natural ability to leverage both the functional information (nodes input features) and structural information (graph adjacency matrix) of graph data using message passing and message aggregation operations. 

The success of GNNs in addressing the graph classification task nevertheless
has been contingent on the availability of a large set of labeled graph samples, 
which induces a significant annotation 
burden in many domains where the labels are scarce and 
require substantial domain-expertise to generate. 
To tackle this problem, 
various graph augmentation methods have been proposed to
increase the size and diversity of the training set by generating additional new graph samples. 
Most common graph augmentation methods, 
such as DropEdge \cite{Rong2020DropEdge}, DropNode \cite{you2020graph}, and SoftEdge \cite{guo2022softedge}, 
involve perturbing the nodes, edges, or subgraphs of a given graph sample to generate a new graph. 
However such methods assume that the employed graph-augmentation operations are label invariant, 
which is difficult to guarantee 
in many cases. Additionally, these methods use a single graph sample to generate new graph instances, which limits the diversity of the generated graphs. Although mixup-based augmentation methods have demonstrated tremendous success 
in improving the generalization capacity
of deep neural networks on image-based \cite{zhang2018mixup} and text-based tasks \cite{sun2020mixup}, it remains an open challenge to apply mixup to graph-based tasks given the irregular, discrete and not well-aligned nature of graph data. 
Few works have proposed methods 
to adapt mixup to graph data, including 
G-Mixup \cite{han2022g} and M-Mixup \cite{wang2021mixup}. However, these methods either are computationally expensive and need a relatively large number of graph samples to obtain good performance 
or generate new graph samples in the manifold space and offer
limited improvement in performance. 

In this work, we propose a novel 
mixup-based graph augmentation method named Graph Dual Mixup (GDM) 
for graph classification, 
which
applies parallel mixup to the functional and structural information of the graph samples to generate new graph instances in the input space. Given the discrete nature of the graph structures, GDM employs a Graph Structural Auto-Encoder (GSAE) to learn a structural embedding of the 
graph nodes. 
It then applies mixup to the learned structural node embeddings of existing graphs to 
generate structural node embeddings for new mixup graph samples,
which are subsequently used to produce the graph structures (i.e., adjacency matrices) of the new graph samples
using the Graph Structural Decoder. 
Regarding 
the functional information, GDM applies mixup directly to the input node features of existing graphs
to obtain the input node features of the corresponding mixup graph samples. 
The new graph instances generated through the parallel mixup over both the input features and graph structures
are thereafter used to supplement the original set of labeled graph samples, reduce overfitting, and help GNNs generalize better 
with scarce graph labels.
Furthermore, 
we propose two Balanced Graph Sampling methods to 
guide the mixup procedure to achieve 
balanced difficulty and diversity
for the generated graph instances. 
We conduct comprehensive experiments on six graph classification benchmark datasets. 
The experimental results demonstrate that our proposed method substantially outperforms state-of-the-art graph augmentation methods in the literature when the number of labeled graphs is limited.

\section{Related Works}

\subsection{Graph Classification}
Earlier works have addressed the graph classification task using graph-kernel based methods where the graph samples are decomposed into small subgraphs \cite{haussler1999convolution,shervashidze2011weisfeiler,yanardag2015deep}. More recently, Graph Neural Networks (GNNs) have been successfully adopted 
in tackling the graph classification task. 
Many
GNN models such as Graph Convolution Networks (GCNs) \cite{kipf2017semisupervised}, Graph Attention Networks (GATs) \cite{velickovic2018graph}, GraphSAGE \cite{hamilton2017inductive} and Graph Isomorphism Networks (GINs) \cite{xu2018how} have been shown to possess strong capacity to represent the graph data using message passing and message aggregation operations, 
and facilitate graph classification. 
Moreover,
some works have developed novel graph readout methods to obtain discriminative graph-level representations from the node-level representation learned by various GNN models \cite{ying2018hierarchical,bianchi2020spectral}. 

\subsection{Graph Augmentation}
Data augmentation methods play a crucial role in regularizing the training of deep models. Common graph augmentation methods are perturbation-based methods that augment graph samples by applying perturbations to graph nodes \cite{you2020graph,huang2018adaptive}, edges \cite{guo2022softedge,Rong2020DropEdge}, or subgraphs \cite{wang2020graphcrop,park2022graph}. DropEdge randomly drops a number of edges from the graph structure during training \cite{Rong2020DropEdge}. 
SoftEdge selects a random subset of edges and assigns 
random weights to them to generate augmented graphs while preserving the connectivity patterns of the input graphs \cite{guo2022softedge}. DropNode randomly deletes a subset of the nodes in the graph together
with their connections to generate augmented graph samples \cite{you2020graph}. 
GraphCrop augments the graphs with sub-structure deletion, which motivates GNNs to learn a robust global-view of the graph samples \cite{wang2020graphcrop}. 
Graph Transplant uses subgraph transplantation to augment graphs where node saliency is used to select the transplanted subgraphs \cite{park2022graph}. 
These
methods however operate under the strong assumption that the applied graph perturbations are label-invariant 
insofar the augmented graph shares the same ground-truth label as the original graph. 
Such an assumption is hard to guarantee in many cases. 
Meanwhile, although there has been 
tremendous success of Mixup-based methods in regularizing deep models in domains where the data is regular, well-aligned and continuous such as images \cite{zhang2018mixup} and text \cite{sun2020mixup}, few works have attempted to adapt mixup to graph data. 
M-Mixup applies mixup to the graph-level representation in the manifold space learned by GNNs  
in a similar way
to manifold mixup \cite{wang2021mixup}. 
G-Mixup performs mixup to the graphons of different classes which are learned from the graph samples, 
and generates augmented graphs
by sampling from the mixed graphons \cite{han2022g}. GraphMix is a node-level augmentation method where manifold mixup is applied to a fully-connected network that is trained jointly with a GNN \cite{verma2021graphmix}. Further details on graph augmentation methods can be found in \cite{zhao2022graph}.

\section{Method}

\subsection{Problem Setup}
We consider the following graph classification setting. 
The input is a set of $N$ labeled graphs: 
$\mathcal{G}=\{(G_1,\mathbf{y}_1),\cdots,(G_N,\mathbf{y}_N) \}$.
Each graph $G$ is made up of a pair $(V, E)$, where $V$ is the set of graph nodes  
with size $|V|= n$ and $E$ is the set of edges. $E$ is represented by an adjacency matrix $A$ of size $n \times n$. 
The adjacency matrix can have either binary or weighted values,  
be symmetric (in the case of undirected graphs) or asymmetric (in the case of directed graphs). 
Each node in the graph $G$ is associated with a corresponding feature vector of size $d$. 
The feature vectors of all the nodes in the graph are represented by an input feature matrix 
$X\in\mathbb{R}^{n\times d}$. 
The graphs in the training set $\mathcal{G}$ may potentially have different sizes (different number of nodes), 
while the feature vectors of the nodes of all graphs have the same size $d$. 
The graph label vector $\mathbf{y}$ is a one-hot label indicator vector of size $C$, where $C$ is the number of classes.

\subsection{Graph Classification}
GNNs address the graph classification task by utilizing both the graph adjacency matrix and the input node features, 
which correspond to the structural and functional information of the graphs, respectively. 
GNN models in the literature are commonly made up of three components: 
a node representation learning function $f_{\theta}$, a graph readout function, and a graph classification function $g_{\phi}$. 
The node representation function $f_{\theta}$ 
typically consists of multiple (e.g., $L$)
GNN layers,
each of which performs message propagation and message aggregation at the node level to learn new node embedding as follows: 
\begin{equation}\label{eq:node_embed}
   \mathbf{h}_u^{l} = \text{AGGREGATE}(\mathbf{h}_{u}^{l-1}, \mathbf{h}_{v}^{l-1} | v \in \mathcal{N}(u), \theta^{l} )
\end{equation}
where $\mathbf{h}_u^{l}\in\mathbb{R}^{d_{l}\times1}$ is the learned embedding of node $u$ with size $d_l$ at layer $l$,  
$\mathcal{N}(u)$ is the set of neighboring nodes of node $u$, 
$\theta^{l}$ is the learnable parameters of the $l$-th GNN layer, 
and $\text{AGGREGATE}$ is the message aggregation function which can be any permutation invariant function 
(sum, average, max, etc.). The initial node embedding $\mathbf{h}_u^{0}$ is the input node feature vector $\mathbf{x}_u$. The graph readout function is a permutation-invariant function used to obtain the graph-level embedding from the learned node-level embedding as follows: 
\begin{equation}\label{eq:pooling}
    \mathbf{h}_G = \text{READOUT}(\mathbf{h}_{u}^{L} | u \in V )
\end{equation}
where $\mathbf{h}_{u}^{L}\in\mathbb{R}^{d_{L}\times1}$ is the embedding of node $u$ obtained from the top layer $L$ of $f_\theta$ and $\mathbf{h}_G\in\mathbb{R}^{d_{G}\times1}$ is the graph-level embedding. The graph classification function $g_{\phi}$ takes the graph-level embedding $\mathbf{h}_G$ as input to 
produce the
predicted class probability vector for the given graph $G$ as follows: 
\begin{equation}\label{eq:graph_class}
 \mathbf{p}_G = g(\mathbf{h}_G|\phi).
\end{equation}
All the components 
are trained end-to-end by minimizing the following cross-entropy loss over the labeled graphs in the training set: 
\begin{equation}\label{eq:loss}
    \mathcal{L} = \sum\nolimits_{G \in \mathcal{G}} \ell(\mathbf{p}_G,\mathbf{y}_G)
\end{equation}
where 
$\ell(\cdot, \cdot)$ is the cross-entropy loss function,
$\mathbf{p}_G$ and $\mathbf{y}_G$ are the predicted class probability vector and the ground-truth label indicator vector 
for graph $G$, respectively.

\subsection{Mixup}

Mixup is an interpolation-based augmentation method that has demonstrated significant success in reducing overfitting and improving the generalization of deep neural networks \cite{zhang2018mixup,sun2020mixup}. Mixup generates augmented training samples $(\tilde{x},\tilde{y})$ by applying linear interpolation between 
a randomly sampled pair of input instances and their corresponding labels as follows: 
\begin{equation}
\begin{split}
\tilde{x} = \lambda x_i + (1-\lambda) x_j, \qquad
\tilde{y} = \lambda {y}_i + (1-\lambda) {y}_j
\end{split}
\end{equation}
where $\lambda$ is a scalar mixing coefficient sampled from a Beta distribution $\text{Beta}(\alpha,\beta)$ with hyper-parameters $\alpha$ and $\beta$. $(\tilde{x},\tilde{y})$ is the new sample generated by mixing the input labeled samples $(x_i, {y}_i)$ and $(x_j, {y}_j)$. Mixup can be readily applied to any classification task where the input data is regular, continuous and well-aligned such as images, text and time-series data. However, mixup cannot be applied directly to graph data given that: 
(1) graph data is irregular where different graphs may potentially have different sizes (different number of nodes). 
(2) graphs do not have a natural-ordering of their nodes, therefore aligning a pair of graphs is a non-trivial task. 
(3) graph structures may be discrete where the edges are binary whereas mixup generates continuous samples. Therefore, it is important to develop new methods that adapt mixup to the discrete, irregular and not well-aligned graph data.

\subsection{Graph Dual Mixup}

In this section, we introduce our proposed Graph Dual Mixup (GDM) method which 
generates new graph samples by applying parallel structural (i.e., structure-based) mixup and functional (i.e., feature-based) mixup  
over each selected pair of existing graph samples. 
In particular, GDM employs a Graph Structural Auto-Encoder (GSAE) to learn a structural embedding of the graph nodes based on the adjacency matrix. 
The structural mixup is then applied on the structural node embeddings of the input pair of graphs to produce a new set of node embeddings, 
which is used to generate the adjacency matrix (i.e., graph structure) of the mixup graph sample using the Graph Structural Decoder of the GSAE. 
As for the functional information encoded with node features, 
GDM applies mixup directly to the input node features to obtain the node features of the generated 
mixup graph sample. 
In the remainder of this section, we elaborate on the dual mixup procedure of this GDM methodology.

\subsubsection{Structural Graph Node Representation Learning}

Given the discrete
nature of graph structures,  
mixup cannot be directly  
applied to the structures of a pair of graphs (represented by their corresponding adjacency matrices) to generate a new graph structure. 
Therefore, we propose to employ a Graph Structural Auto-Encoder (GSAE) to learn a structural embedding of the graph nodes
and support mixup 
in the learned structural embedding space. 
This allows us to evade the difficulties associated with applying mixup to the original graph structures. 
GSAE
is made up of a structural encoder $\mathcal{E}_s$  
and a structural decoder $\mathcal{D}_s$.  
The structural encoder $\mathcal{E}_s$ consists of multiple GNN layers that learn the structural node embeddings by propagating and aggregating messages 
across the graph structure, where the messages reflect solely the structural information of the nodes. 
The goal is to learn a structural embedding of all the nodes in the graph that would enable us to reconstruct the graph adjacency matrix. 
Specifically, for a given graph sample $G=(X,A)$, 
$\mathcal{E}_s$ takes 
the adjacency matrix $A\in\mathbb{R}^{n\times n}$ and the 
node degree matrix $D\in\mathbb{R}^{n\times n}$ (represent the initial node structural features) computed from $A$ as input 
 to learn the structural node embeddings as follows: 
\begin{equation}\label{eq:strct_enc}
	H_{s} =  \mathcal{E}_{{s}}(D, A), \qquad  \text{where} \, \, D[i,i] = \sum\nolimits_j A[i,j], 
\end{equation}
where the node degree matrix $D$ is an identity matrix whose main diagonal values correspond to the degrees of the associated nodes; 
$H_{s}\in\mathbb{R}^{n\times d_s}$ is the learned structural embedding of the nodes in the graph with size $d_s$.
$H_s$ holds solely the structural information of all the nodes in the graph,
from which  
one can 
reconstruct the connections/edges between the nodes 
and therefore 
the original adjacency matrix $A$ 
using the structural decoder $\mathcal{D}_s$ of the GSAE. 
In particular, we adopt a simple inner product similarity based decoder as the  structural decoder $\mathcal{D}_s$, which 
takes the learned structural node embeddings as input to reconstruct the graph adjacency matrix $A$ as follows: 
\begin{equation}\label{eq:strct_dec}
	\hat{A} = \mathcal{D}_s( H_{s}) = \sigma( H_{s} H^{T}_{s})
\end{equation}
where $\sigma$ is the sigmoid activation function and $\hat{A}\in\mathbb{R}^{n\times n}$ is the decoded/re-constructed adjacency matrix. The GSAE is trained end-to-end to minimize the following graph structure reconstruction loss:

\begin{equation}\label{eq:SGAE_loss}
               \mathcal{L}^{s}_{\text{re}}  =   - \sum\nolimits_{G \in \mathcal{G} } \Big[ \sum\nolimits_{(i,j) \in E_G} \log( \hat{A}_G[i,j])
	       + \sum\nolimits_{(i,j) \in S_{G}^{\text{neg}}} \log(1- \hat{A}_G[i,j])   \Big]
\end{equation}
where $E_G$ is the set of edges for graph $G$ and $S_{G}^{\text{neg}}$ is the set of randomly sampled negative edges of graph $G$ (i.e. edges that do not exist in the original graph). It is important to note that GSAE does not access/use the input node features (functional graph information) as it replaces the input node features with the corresponding node degrees calculated from the adjacency matrix. GSAE also does not make use of the graph class labels as it is learned in a completely self-supervised/unsupervised fashion. 

\subsubsection{Graph Generation via Dual Mixup}

After training the GSAE, our proposed Graph Dual Mixup is ready to apply Structural Mixup and Functional Mixup to the structural and functional information of the graphs respectively to generate new graph samples. 
To achieve that, for a given pair of graphs and their corresponding label vectors $(G_i,\mathbf{y}_i)$ and $(G_j,\mathbf{y}_j)$, 
where the two graphs are made up of input node feature matrices and graph adjacency matrices 
such as $G_i = (X_i,A_i)$ and $G_j = (X_j,A_j)$, 
GDM randomly aligns the nodes of the graph pair. 
When $G_i$ and $G_j$ have different sizes ($n_i \neq n_j$), we pad the input node feature matrix and adjacency matrix of the smaller graph with zeros to match the size 
of the larger graph. 
Then we apply functional mixup directly to the input node features and the label vectors of the graph pair to generate the node features
of the new graph sample $\tilde{G}$ and its corresponding label vector $\tilde{\mathbf{y}}$ as follows:
\begin{equation}\label{eq:GDM_x_y}
\begin{split}
\tilde{X} = \lambda  X_i  + (1-\lambda) X_j, \qquad %
\tilde{\mathbf{y}} = \lambda  \mathbf{y}_i  + (1-\lambda) \mathbf{y}_j
\end{split}
\end{equation}

To obtain
the structural information of the generated new graph sample $\tilde{G}$, 
GDM applies structural mixup in the structural embedding space learned by the GSAE as follows: 
\begin{equation}\label{eq:GDM_Hs}
	\tilde{H}_s = \lambda  \, \mathcal{E}_{{s}}(D_i ,A_i)  + (1-\lambda) \,  \mathcal{E}_{{s}}(D_j ,A_j)  
\end{equation}
where $D_i$ and $D_j$ are the degree matrices of  
$G_i$ and $G_j$, respectively; 
$\tilde{H}_s\in\mathbb{R}^{\max(n_i,n_j)\times d_s}$ is the structural node embedding matrix of the generated graph $\tilde{G}$. 
The graph structural decoder is then used to reconstruct the adjacency matrix of graph $\tilde{G}$ from the mixed structural node embeddings: 
\begin{equation}\label{eq:GDM_A}
	\tilde{A}  = \mathcal{D}_{{s}}(\tilde{H}_{s})=\sigma(\tilde{H}_{s}\tilde{H}_s^T) 
\end{equation}
The obtained 
matrix $\tilde{A}\in\mathbb{R}^{\max(n_i,n_j)\times \max(n_i,n_j)}$ is a weighted adjacency matrix with edge weights between 0 and 1. 
In order to filter out the noise in the edge weights  
and sparsify the structure of generated graph sample, we prune 
the adjacency matrix by dropping off the
weak edges with 
weights smaller than a pre-defined threshold $\epsilon$ as follows:
\begin{equation}\label{eq:GDM_prune_A}
    \tilde{A}[i,j]= \begin{cases}
    \tilde{A}[i,j],& \text{if } \, \tilde{A}[i,j]\geq \epsilon \\
    0,              & \text{otherwise}.
\end{cases}
\end{equation}
Moreover, 
in order for the structure of the generated graph sample $\tilde{G}$ to match the structural properties of the original graph samples, we post-process $\tilde{A}$ accordingly. In the case that the original graph samples 
have weighted edges,
no post-processing is required. As for the case of the original graph samples being unweighted/binary graphs, we binarize $\tilde{A}$ by replacing all its non-zero values with value 1 as follows: 

\begin{equation}\label{eq:GDM_binarize_A}
    \tilde{A}[i,j]= \begin{cases}
    1,& \text{if } \, \tilde{A}[i,j] > 0 \\
    0,              & \text{otherwise}
\end{cases}
\end{equation}
In this manner, we obtain a new generated graph $\tilde{G}$ with its mixup node features $\tilde{X}$, adjacency matrix $\tilde{A}$ and label vector $\tilde{\mathbf{y}}$.

\subsection{Balanced Graph Sampling}

Given the limited number of available labeled graph instances, randomly sampling pairs of graphs to generate new graph instances 
might be inadequate for
improving model generalization and reducing overfitting 
as random sampling does not take the difficulty or diversity of the generated graph instances into consideration. 
Therefore, we propose two novel Balanced Graph Sampling methods 
to enhance
the diversity and balanced difficulty of the generated graph samples. 
The proposed methods can separately: 
(1) generate low difficulty graphs by applying GDM to randomly sampled pairs of low difficulty graphs; 
(2) generate medium difficulty graphs by applying GDM to mix randomly sampled low difficulty graphs with high difficulty graphs; 
and (3) generate high difficulty graphs by applying GDM to randomly sampled pairs of high difficulty graphs.
The advantage of balanced graph sampling over random sampling is that it guarantees that
the generated graph samples have 3 subsets with \textit{equal sizes}:
a low difficulty subset, a medium difficulty subset, and a high difficulty subset.

To achieve that, we need to assess/estimate the difficulty level of the original graph instances. 
This
is accomplished by pre-training a 
GNN model on the original set of labeled graph instances to minimize the classification loss 
shown in Eq.(\ref{eq:node_embed})---Eq.(\ref{eq:loss}). 
Then the pre-trained GNN model is used to evaluate the difficulty level of each graph $G$
based on its  
predicted class probability vector $\mathbf{p}_G$. 
The first balanced graph sampling method is an Accuracy-based method (Acc), 
which determines the level of difficulty for graph $G$ based on 
 the accuracy/correctness of its predicted class label:
\begin{equation}\label{eq:GDM_Acc}
   \text{Diff}_{\text{Acc}}(G) = \begin{cases}
    \text{low},& \text{if } \, \text{argmax}\; \mathbf{p}_G = \text{argmax}\; \mathbf{y}_G  \\
    \text{high},              & \text{otherwise}
\end{cases}
\end{equation}

The second balanced graph sampling method is an Uncertainty-based method (Unc), 
which uses the uncertainty/entropy of the model prediction on a sample graph $G$ to determine its level of difficulty. 
In particular, we sort the graphs from the training set $\mathcal{G}$ based on the entropy of their corresponding predicted class probability vectors, 
then consider 
the graphs with the lowest half of entropy scores to be low difficulty ones while the other half of the graphs are taken as
high difficulty ones: 

\begin{equation}\label{eq:GDM_Unc}
   \text{Diff}_{\text{Unc}}(G) = \begin{cases}
    \text{low},& \text{if } \, \text{Ent}(\mathbf{p}_G) \leq  \text{Med}\bigl(\{\text{Ent}(\mathbf{p}_1), \cdots ,\text{Ent}(\mathbf{p}_N)\}\bigr)\\
    \text{high},              & \text{otherwise}
\end{cases}
\end{equation}
where $\text{Ent}(.)$ is the entropy function and $\text{Med}(.)$ is the median function. 
Therefore, GDM can be applied with Accuracy-based Balanced Graph Sampling (GDM Acc) or Uncertainty-based Balanced Graph Sampling (GDM Unc) 
to generate a new set of diverse graph samples $\mathcal{G}_{\text{GDM}}$ with balanced difficulty. 

\subsection{Augmented Training Procedure}

The combination of Balanced Graph Sampling and Graph Dual Mixup generates a diverse set of new graph instances, 
which can supplement the limited number of original labeled graph samples. 
Finally, we train the GNN model using the original graph set $\mathcal{G}$ and the generated graph set $\mathcal{G}_{\text{GDM}}$ by minimizing the following loss function:
\begin{equation}\label{eq:total_loss}
     \mathcal{L}_{\text{total}}  =  \sum_{G \in \mathcal{G}}  \ell_{CE}(\mathbf{p}_G,\mathbf{y}_G) +   \lambda_{\text{GDM}}  \sum_{G \in 
 \mathcal{G}_{\text{GDM}}} \ell_{CE}(\mathbf{p}_G,\tilde{\mathbf{y}}_G)  
\end{equation}
where $\lambda_{\text{GDM}}$ is a trade-off hyper-parameter controlling the contribution of the generated graph set $G_{GDM}$. 
An overview of the graph augmentation process and the GNN augmented training procedure is presented in Algorithm \ref{alg:method}.

\begin{algorithm}[t]\normalsize
    \caption{Augmentation and Training Procedure of Graph Dual Mixup}
    \label{alg:method}
    \begin{algorithmic}
\STATE{\textbf{Input:} Graph set $\mathcal{G}$; hyper-parameters $\alpha$, $\beta$ $\epsilon$, $\lambda_{\text{GDM}}$ }
\STATE{\textbf{Output:} Learned model parameters $\theta$, $\phi$ }
    \STATE Pre-train a GNN Model on $\mathcal{G}$ 
	    to determine the graph difficulty levels
	    \STATE  Train GSAE on $\mathcal{G}$ using Eq.(\ref{eq:strct_enc}), (\ref{eq:strct_dec}), (\ref{eq:SGAE_loss}). 
\STATE $\mathcal{G}_{\text{low}}  =$ Generate low difficulty samples with GDM 
\STATE $\mathcal{G}_{\text{med}} =$ Generate medium difficulty samples with GDM 
\STATE $\mathcal{G}_{\text{high}} =$ Generate high difficulty samples with GDM 
\STATE $\mathcal{G}_{\text{GDM}} =  \mathcal{G}_{\text{low}} \cup \mathcal{G}_{\text{med}} \cup\mathcal{G}_{\text{high}} $  
	    \STATE Train the final GNN Model on $\mathcal{G}$ and $\mathcal{G}_{\text{GDM}}$ using Eq.(\ref{eq:node_embed}), (\ref{eq:pooling}), (\ref{eq:graph_class}), and (\ref{eq:total_loss}).
    \end{algorithmic}
\end{algorithm}

\section{Experiments}

\subsection{Experimental Setup}

\subsubsection{Datasets \& Baselines}
We evaluate our proposed method on 6 graph classification benchmark datasets from the TUDatasets \cite{Morris2020}, including 3 chemical datasets and 3 social datasets. The chemical datasets are D\&D \cite{dobson2003distinguishing},
Proteins \cite{borgwardt2005protein} and NCI1 \cite{wale2008comparison}, while the social datasets are IMDB-Binary, IMDB-Multi and Reddit-5K \cite{yanardag2015deep}. 
We employ the same 10-fold train/validation/test split provided by \cite{erricafair}. 
We apply our proposed Graph Dual Mixup on the Graph Convolution Network (GCN) baseline \cite{kipf2017semisupervised}
and compare our proposed method against 5 other graph augmentation methods from the literature: DropNode \cite{you2020graph}, DropEdge \cite{Rong2020DropEdge}, M-Mixup \cite{wang2021mixup}, SoftEdge \cite{guo2022softedge} and G-Mixup \cite{han2022g}.

\begin{table}[t]\normalsize
\centering
\caption{Mean classification accuracy (standard deviation is within brackets) on 6 graph classification benchmark datasets with 10 labeled graphs per class.}
\begin{tabular}{l|llllll}
\hline
Dataset  & \multicolumn{1}{c}{Proteins} & \multicolumn{1}{c}{NCI1} & \multicolumn{1}{c}{D\&D} & \multicolumn{1}{c}{IMDB-B} & \multicolumn{1}{c}{IMDB-M} & \multicolumn{1}{c}{Reddit} \\
\hline
GCN      & $59.3_{(6.8)}$               & $51.0_{(1.6)}$           & $59.5_{(2.7)}$         & $54.5_{(3.9)}$             & $36.9_{(3.7)}$             & $25.1_{(5.1)}$             \\
DropNode & $61.0_{(8.5)}$               & $52.9_{(3.4)}$           & $62.1_{(2.9)}$         & $59.0_{(5.7)}$             & $36.9_{(4.6)}$             & $30.8_{(8.4)}$             \\
DropEdge & $59.4_{(5.8)}$               & $53.1_{(3.7)}$           & $62.6_{(4.5)}$         & $57.6_{(5.5)}$             & $37.2_{(4.1)}$             & $26.7_{(8.4)}$             \\
SoftEdge & $58.9_{(7.2)}$               & $52.0_{(3.2)}$           & $59.5_{(2.4)}$         & $55.3_{(6.6)}$             & $36.2_{(3.0)}$             & $25.0_{(4.9)}$             \\
M-Mixup  & $59.0_{(7.2)}$               & $51.9_{(3.3)}$           & $59.1_{(5.3)}$         & $57.1_{(6.4)}$             & $37.4_{(5.3)}$             & $23.0_{(2.8)}$             \\
G-Mixup  & $60.8_{(2.1)}$               & $51.8_{(3.2)}$           & $58.7_{(4.2)}$         & $55.1_{(8.5)}$             & $36.9_{(4.3)}$             & $24.1_{(7.3)}$             \\
GDM Acc & $\mathbf{66.0}_{(5.3)}$               & $\mathbf{57.5}_{(2.6)}$           & $62.1_{(3.7)}$         & $\mathbf{61.3}_{(6.7)}$             & $\mathbf{40.9}_{(5.4)}$             & $\mathbf{36.3}_{(8.0)}$             \\
GDM Unc & $65.1_{(6.1)}$               & $56.8_{(3.9)}$           & $\mathbf{64.0}_{(4.2)}$         & $61.0_{(7.0)}$             & $39.8_{(5.5)}$             & $34.9_{(9.1)}$ \\
\hline           
\end{tabular}
\label{table:10labelrate}
\end{table}

\begin{figure}[ht]
\centering
\includegraphics[width=0.95\textwidth]{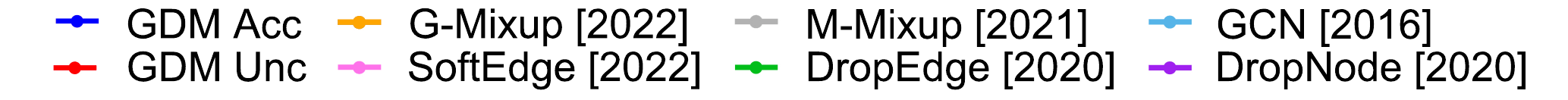}
\begin{subfigure}{0.32\textwidth}
\centering
\includegraphics[width = \textwidth]{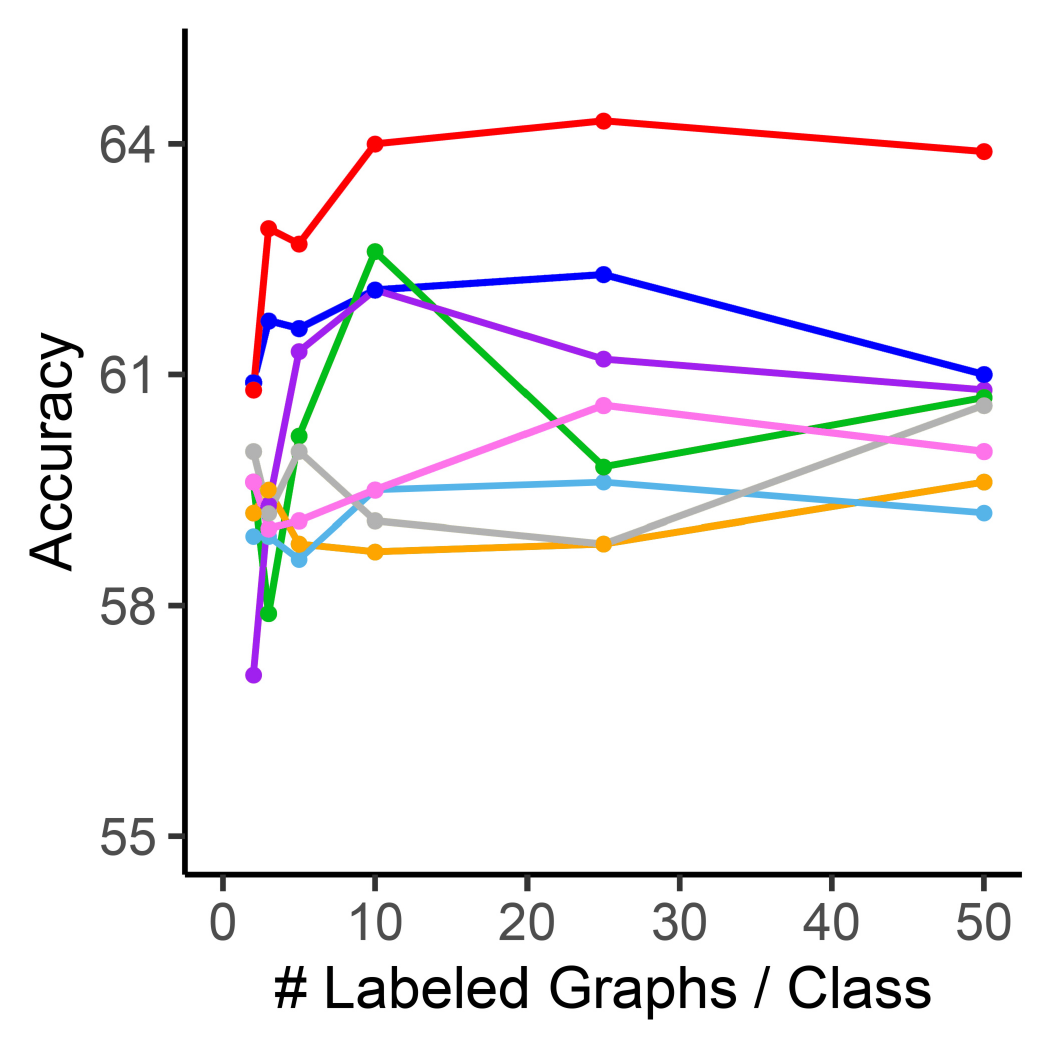} 
\caption{D\&D}
\end{subfigure}
\begin{subfigure}{0.32\textwidth}
\centering
\includegraphics[width = \textwidth]{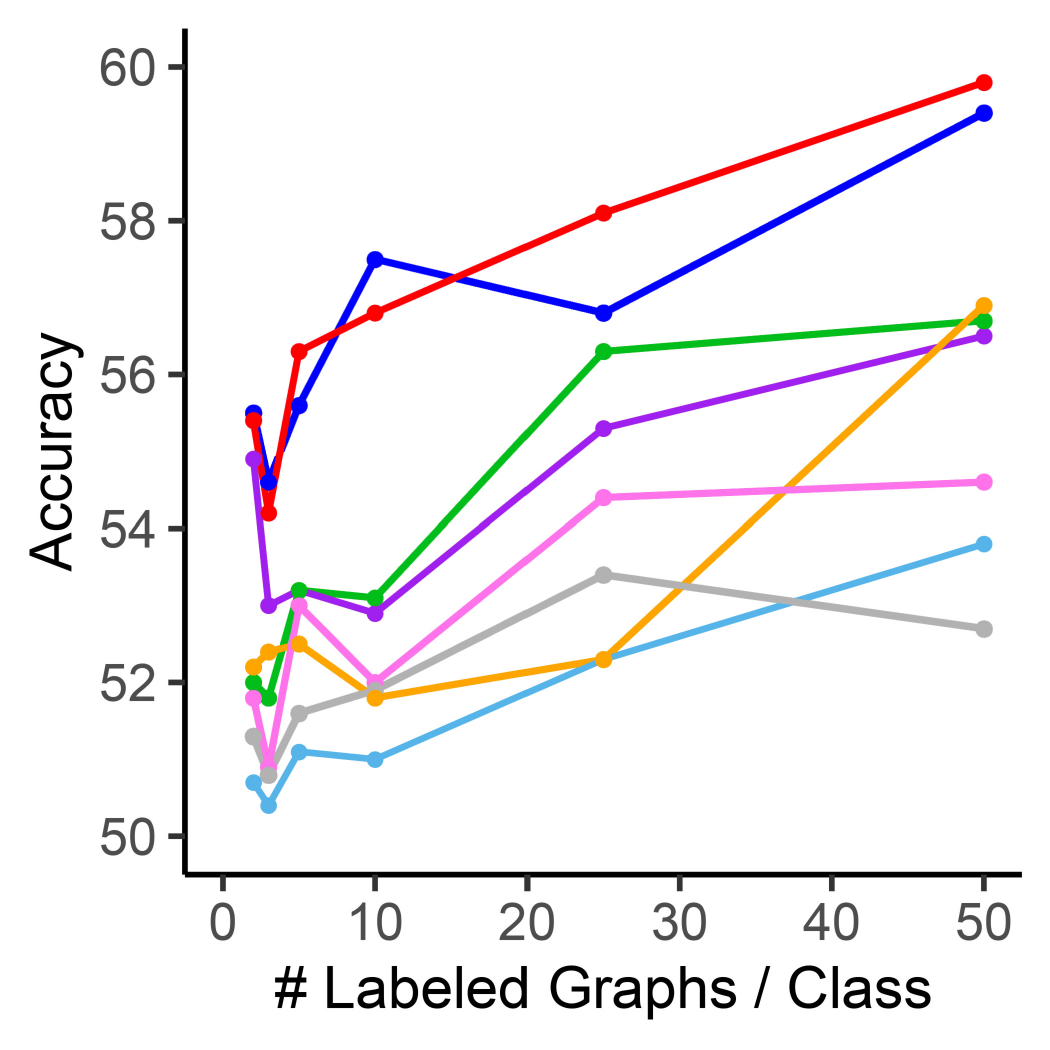}
\caption{NCI1}
\end{subfigure}
\begin{subfigure}{0.32\textwidth}
\centering
\includegraphics[width = \textwidth]{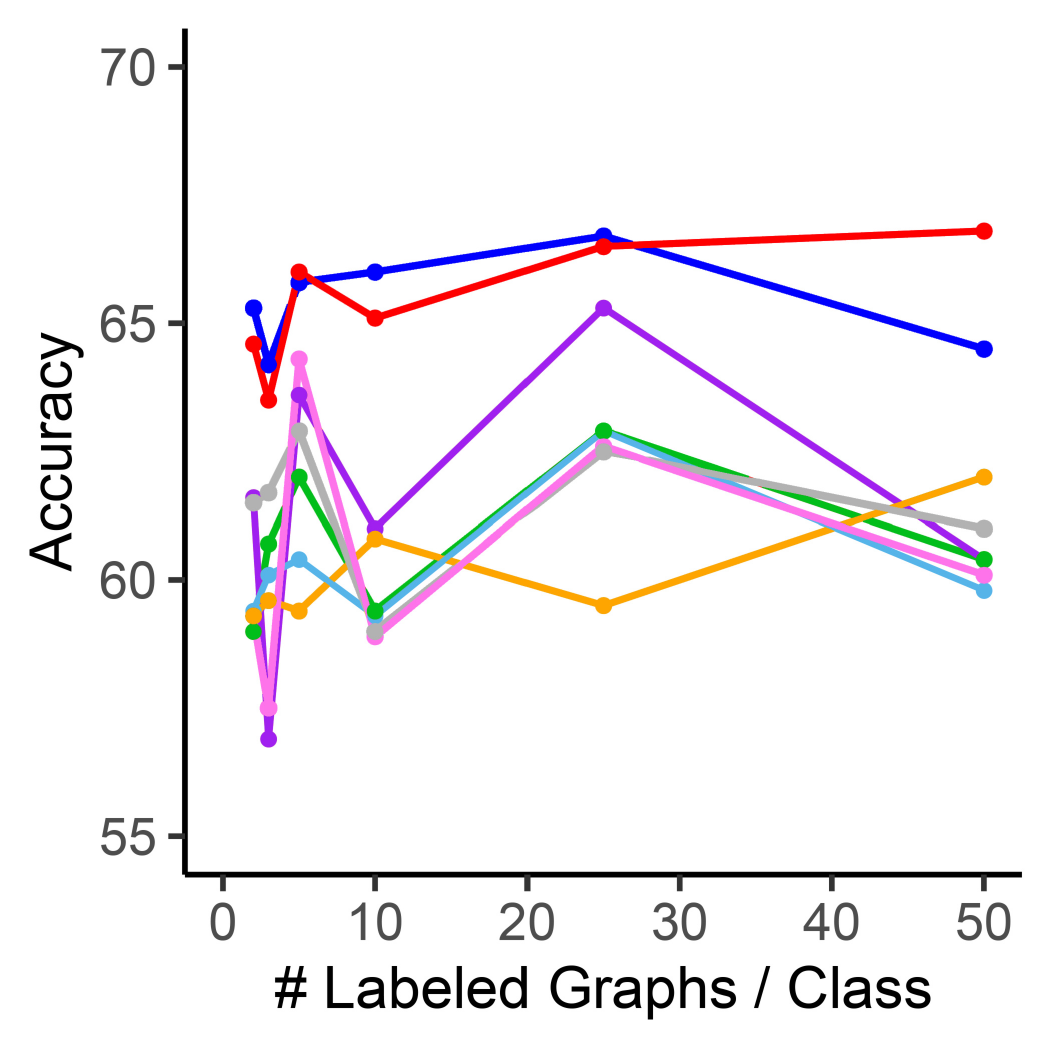}
\caption{Proteins}
\end{subfigure}
\begin{subfigure}{0.32\textwidth}
\centering
\includegraphics[width = \textwidth]{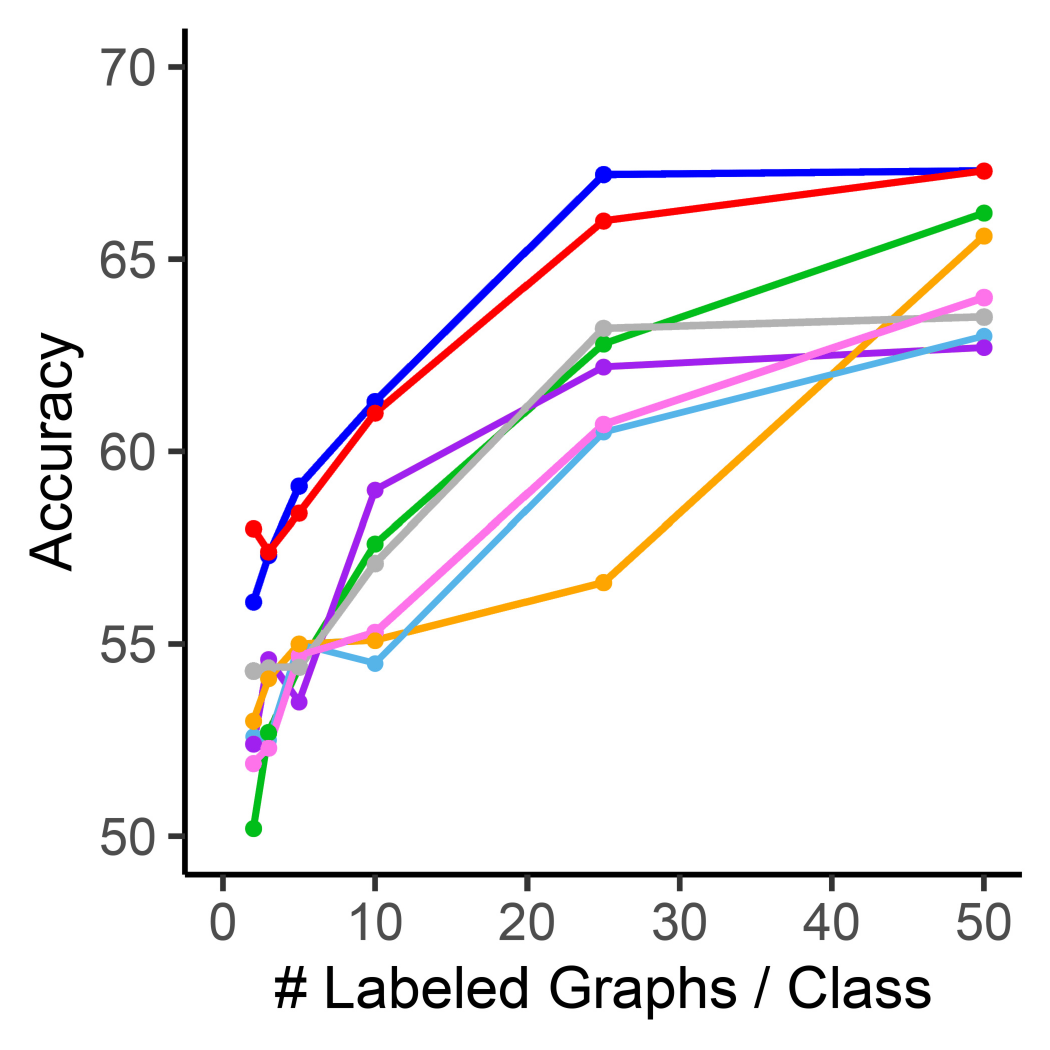} 
\caption{IMDB-Binary}
\end{subfigure}
\begin{subfigure}{0.32\textwidth}
\centering
\includegraphics[width = \textwidth]{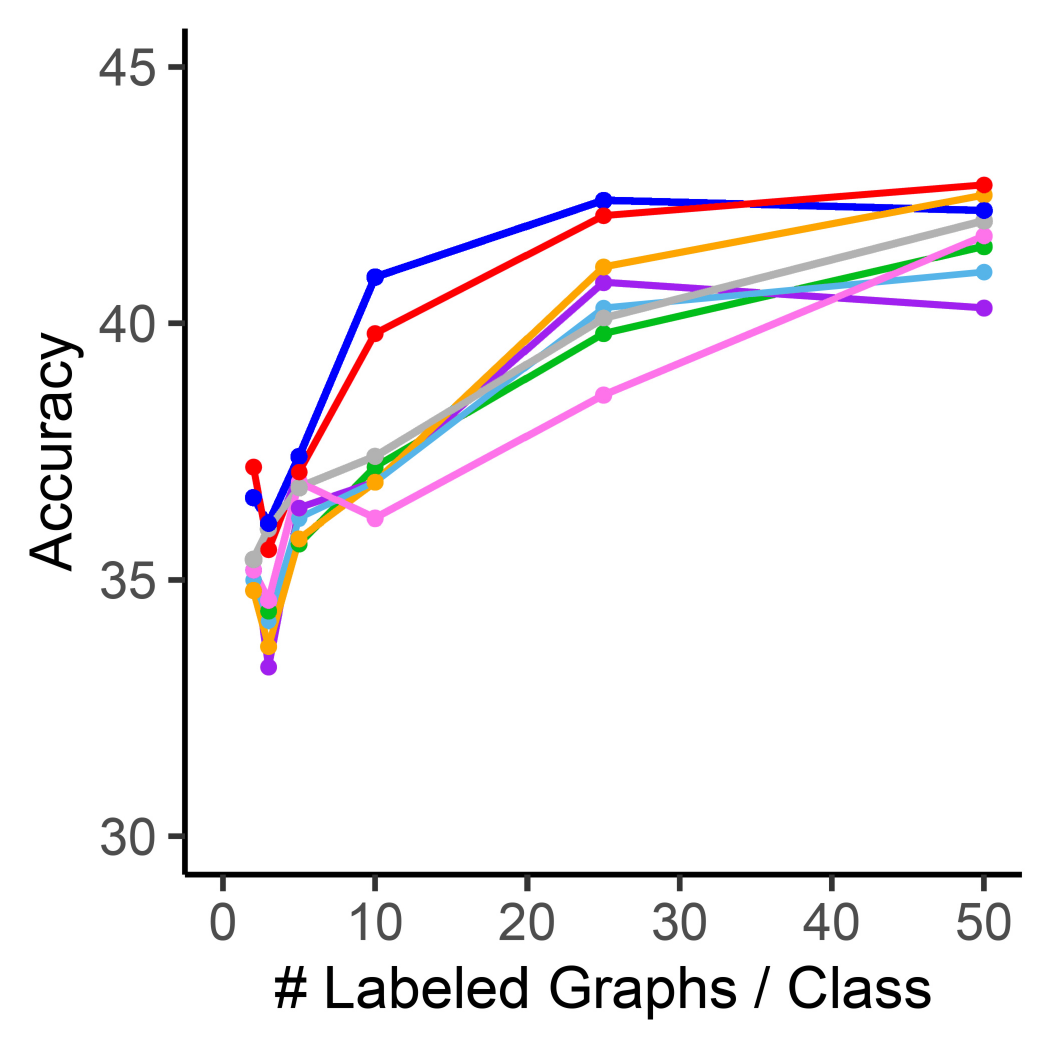}
\caption{IMDB-Multi}
\end{subfigure}
\begin{subfigure}{0.32\textwidth}
\centering
\includegraphics[width = \textwidth]{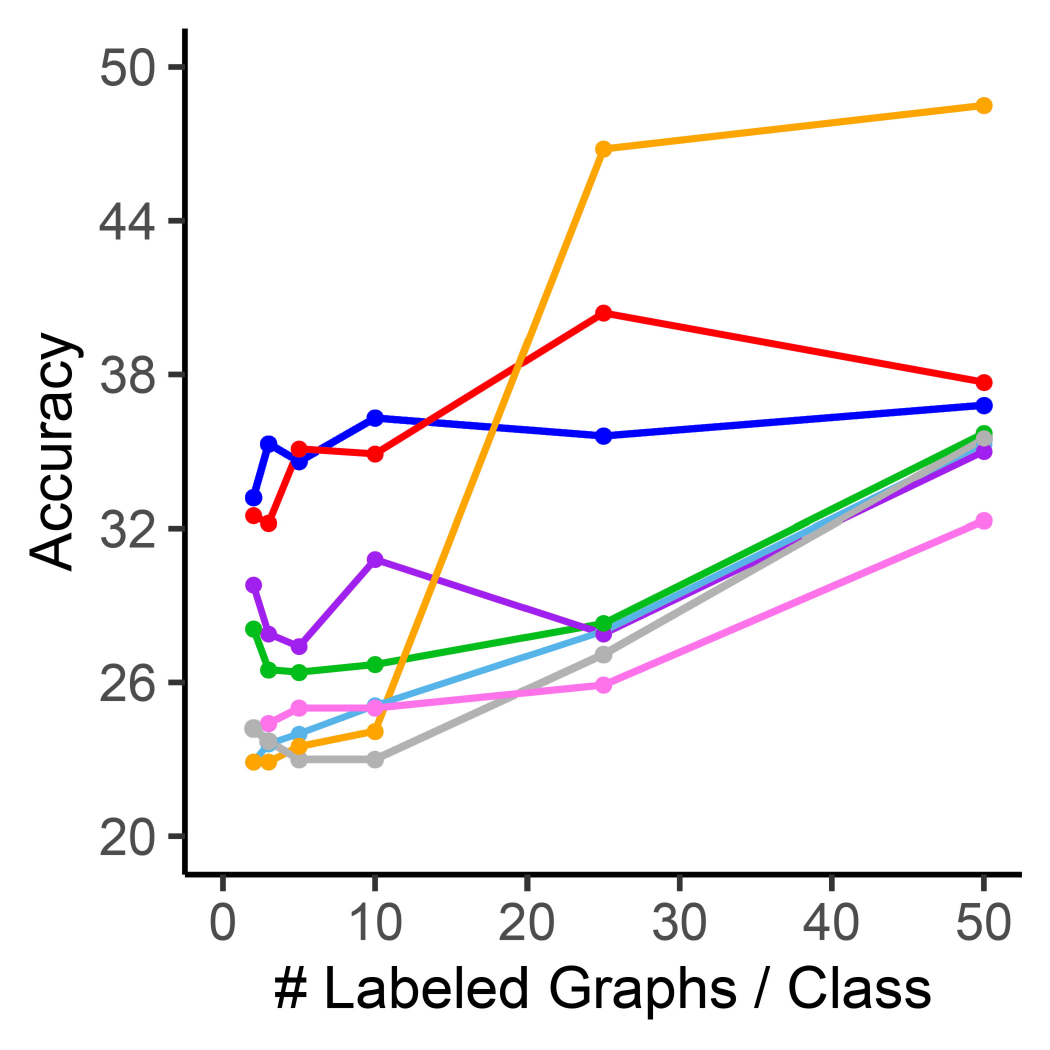}
\caption{Reddit-5K}
\end{subfigure}
\caption{Mean classification accuracy on 6 graph classification benchmark datasets with few labeled graphs per class (2, 3, 5, 10, 25, 50). }
\label{fig:main_exp}
\end{figure}

\noindent\paragraph{\bf Implementation Details}
The node representation function $f_\theta$ of the GNN model is made up of 
4 message passing layers, followed by Global Mean Pooling as the Readout function. 
The graph classification function $g_\phi$ is made up of 2 fully connected layers followed by a softmax function. Each message passing layer and fully connected layer is followed by a Rectified Linear Unit (ReLU) activation function. 
The structural encoder $\mathcal{E}_s$ is made up of 2 GCN message passing layers. The GNN model is pre-trained on the original graph set for 100 epochs and subsequently trained on the original graph set and augmented graph set for 800 epochs, 
both using the Adam optimizer with learning rate of 1e-2. 
The Graph Structural Auto-Encoder is trained for 200 epochs using the Adam optimizer with learning rate of 1e-2. 
The loss trade-off hyperparameter $\lambda_{\text{GDM}}$ and 
the weak edge pruning threshold $\epsilon$ take values 1 and 0.1, respectively. 
The mixing scalar coefficient $\lambda$ is sampled from distribution Beta$(\alpha, \beta)$ with hyper-parameters $\alpha = \beta = 1.0$. We use a dropout rate of 0.25 for SoftEdge, DropNode and DropEdge. 
For G-Mixup, we use the same hyper-parameters reported in \cite{han2022g}.

\subsection{Comparison Results}

We investigate the performance of our proposed GDM with limited numbers of labeled graphs. 
We aim to use a small number of
labeled graphs per class, e.g., \{2, 3, 5, 10, 25, 50\}, as the training set. 
To achieve that, we randomly sampled graphs from the training set of each fold in the 10-fold split provided \cite{erricafair} to match the desired label rates. For each label rate, we repeat our experiments 
3 times 
on all the 10-folds 
and average the test accuracy over all folds and all runs. We evaluate GDM in combination with 
the proposed two Balanced Graph Sampling methods to obtain: 
(1) ``GDM Acc'', where GDM is applied with the Accuracy-based Balanced Graph Sampling; and 
(2) ``GDM Unc'', where GDM is applied with Uncertainty-based Balanced Graph Sampling. We report the obtained test accuracy results with 10 labeled graphs per class in Table \ref{table:10labelrate}, while the test accuracy results for all label rates are presented in Figure \ref{fig:main_exp}.

The results in Table \ref{table:10labelrate} clearly demonstrate that both variants of our proposed GDM greatly
outperform the underlying GCN baseline and the other 5 graph augmentation methods across all 6 datasets. 
GDM improves the performance of the underlying GCN baseline by 6.7\%, 7.5\%, 6.8\% and 11.2\% on the Proteins, NCI1, IMDB-Binary and Reddit-5K datasets, respectively. 
The performance gain over the other graph augmentation methods is also notable, exceeding 5\%, 4.4\% and 6.3\% on Proteins, NCI1 and Reddit-5K, respectively. 
Moreover, Figure \ref{fig:main_exp} clearly shows that our proposed GDM consistently 
outperforms the GCN baseline and the 5 comparison 
graph augmentation methods on 5 datasets across almost all label rates. 
Only in the case of the Reddit-5K dataset with label rates of larger than 25 labeled graphs per class, G-Mixup outperforms our proposed method. Nevertheless, GDM consistently improves the performance of the underlying GCN baseline across all the label rates on all the datasets, 
achieving performance gains over
6\%, 5\%, 5\% and 11\% on Proteins, NCI1, IMDB-Binary and Reddit-5K, respectively, 
in the case of 2 labeled graphs per class. 
Furthermore, GDM yields remarkable
performance gains over the other graph augmentation methods, exceeding 4\% on Proteins, Reddit-5K and IMDB-Binary in the case of 2 labeled graphs per class. This highlights the superior performance of the proposed GDM over the existing state-of-the-art graph augmentation methods 
for graph classification with limited supervision.

\subsection{Ablation Study}
\subsubsection{Impact of Balanced Graph Sampling}
We conduct an ablation study to investigate the impact of our balanced graph sampling methods on the proposed GDM method. Specifically, we consider four variants of the balanced graph sampling: (1) $\text{w/o Low Diff}$: we do not generate low difficulty samples. 
(2) $\text{w/o Med Diff}$: we do not generate medium difficulty samples. 
(3) $\text{w/o High Diff}$: we do not generate high difficulty samples. 
(4) GDM Rand: we drop the proposed balanced graph sampling method and use random sampling for mixup.  
We evaluate the first three variants  
using 
both the GDM Acc and GDM Unc methods of balanced graph sampling. 
The comparison results with different label rates---\{2, 3, 5, 10\} labeled graphs per class---%
on the D\&D and IMDB-Multi datasets are reported in Table \ref{table:Match}. 

From Table \ref{table:Match}, we can see that all variants have a performance drop from the full balanced graph sampling  
on both datasets with almost all label rates for both GDM Acc and GDM Unc. 
The $\text{w/o Low Diff}$ variant produces the most notable performance degradation,
which can be attributed to the GNN models' needs for low difficulty and confident samples to improve generalization and 
prevent underfitting when learning with very low label rates. 
The $\text{w/o Med Diff}$ and $\text{w/o High Diff}$ variants also suffer performance degradations, 
indicating the importance of medium difficulty and high difficulty samples 
for inducing
better generalization and reducing overfitting. Additionally, the GDM Rand variant also
demonstrates notable performance drops
compared to both GDM Acc and GDM Unc with almost all label rates, which highlights the importance of ensuring the diversity and balanced difficulty of the generated graph samples. 
These results validate
the contribution of each component 
in balanced graph sampling.

\begin{table}[t]\normalsize
\caption{Ablation study results on the impact of Balanced Graph Sampling in terms of mean classification accuracy (standard deviation is within brackets) with a few labeled graphs per class  (2, 3, 5, 10).}
\resizebox{\textwidth}{!}{
\begin{tabular}{l|llll||llll}
\hline
   & \multicolumn{4}{c||}{D\&D}                                                                                                                     & \multicolumn{4}{c}{IMDB-Multi}                                                                                                                         \\
 & \multicolumn{1}{c}{2} & \multicolumn{1}{c}{3} & \multicolumn{1}{c}{5} & \multicolumn{1}{c||}{10}   & \multicolumn{1}{c}{2} & \multicolumn{1}{c}{3} & \multicolumn{1}{c}{5} & \multicolumn{1}{c}{10}   \\
\hline
GDM Rand  & ${59.6}_{(5.3)}$        & ${61.2}_{(3.9)}$        & ${61.5}_{(3.5)}$        & ${63.0}_{(4.5)}$  & ${35.4}_{(3.8)}$        & ${35.5}_{(4.0)}$        & $36.4_{(5.5)}$        & ${39.7}_{(5.3)}$               \\
\hline
GDM Acc                            & $\mathbf{61.0}_{(2.6)}$        & $\mathbf{61.7}_{(3.3)}$        & $\mathbf{61.6}_{(3.5)}$        & ${62.1}_{(3.7)}$                   & $\mathbf{36.6}_{(4.9)}$        & $\mathbf{36.1}_{(3.8)}$        & $37.4_{(4.5)}$        & $\mathbf{40.9}_{(5.4)}$               \\

w/o Low Diff & $59.2_{(1.4)}$        & $57.7_{(5.7)}$        & $59.2_{(2.1)}$        & $58.2_{(4.0)}$                       & $35.0_{(3.6)}$        & $34.4_{(2.8)}$        & $34.5_{(1.8)}$        & $36.5_{(3.1)}$              \\

w/o Med Diff & $60.2_{(3.9)}$        & $61.6_{(4.1)}$        & $59.9_{(2.1)}$        & $59.6_{(3.1)}$                      & $34.9_{(3.4)}$        & $35.4_{(4.7)}$        & $38.3_{(4.0)}$        & $39.9_{(5.8)}$                    \\

w/o High Diff & $60.1_{(3.2)}$        & $60.6_{(4.4)}$        & $60.6_{(2.7)}$        & $61.3_{(4.1)}$                       & $33.8_{(3.2)}$        & $34.4_{(3.9)}$        & $\mathbf{38.4}_{(4.3)}$        & $39.0_{(5.5)}$                    \\
\hline
GDM Unc                            & $\mathbf{60.8}_{(2.8)}$        & $\mathbf{62.9}_{(3.7)}$        & $\mathbf{62.7}_{(3.2)}$        & $\mathbf{64.0}_{(4.2)}$             & $\mathbf{37.2}_{(4.9)}$        & $\mathbf{35.6}_{(3.6)}$        & $\mathbf{37.1}_{(5.4)}$        & $\mathbf{39.8}_{(5.5)}$                       \\
  
w/o Low Diff & $59.3_{(1.0)}$        & $58.8_{(1.0)}$        & $58.9_{(1.0)}$        & $59.7_{(1.8)}$                     & $35.8_{(2.9)}$        & $34.7_{(3.7)}$        & $34.4_{(2.7)}$        & $37.0_{(3.4)}$                     \\

w/o Med Diff & $\mathbf{60.8}_{(2.9)}$        & $60.4_{(4.3)}$        & $\mathbf{62.7}_{(3.3)}$        & $62.8_{(4.5)}$                   & $35.3_{(3.6)}$        & $34.8_{(3.4)}$        & $36.6_{(4.9)}$        & $38.8_{(4.6)}$                  \\

w/o High Diff & $60.4_{(2.6)}$        & $61.5_{(3.7)}$        & $62.0_{(4.2)}$        & $63.2_{(3.4)}$                & $35.7_{(3.2)}$        & $35.0_{(3.1)}$        & $\mathbf{37.1}_{(4.0)}$        & $39.6_{(3.8)}$                 \\
\hline
\end{tabular}}
\label{table:Match}
\end{table}

\subsubsection{Impact of Graph Structural Auto-Encoder}  
We further conduct an ablation study to investigate the impact of the Graph Structural Auto-Encoder on the proposed GDM. 
Specifically, we compare our proposed GSAE with a Variational Graph Structural Auto-Encoder (VGSAE). 
The VGSAE learns the parameters of a Gaussian distribution (mean and variance) to represent the underlying structure of the graph \cite{kipf2016variational}. The comparison results with different label rates on the Proteins and IMDB-Binary datasets are reported in Table \ref{table:GAE}. 
From the table, it is clear that GSAE outperforms VGSAE on both datasets across almost all label rates. 
The performance gain of GSAE decreases as the label rate increases, 
which highlights that VGSAE requires more training samples to obtain good performance. Therefore GSAE is more suitable for the case of learning with limited supervision as it is able to obtain good generalization performance with few samples due to its simple architecture and smaller number of learnable parameters. Nevertheless, the proposed GDM greatly and consistently outperforms the underlying GCN baseline across all different label rates with both GSAE and VGSAE on both datasets.

\begin{table}[t]\normalsize
\caption{Ablation study results on the impact of Graph Structural Auto-encoder in terms of mean classification accuracy (standard deviation is within brackets).}
\resizebox{\textwidth}{!}{
\begin{tabular}{l|llll||llll}
\hline
   & \multicolumn{4}{c||}{Proteins}                                                                                                                     & \multicolumn{4}{c}{IMDB-Binary}                                                                                                                         \\
 & \multicolumn{1}{c}{2} & \multicolumn{1}{c}{3} & \multicolumn{1}{c}{5} & \multicolumn{1}{c||}{10} &  \multicolumn{1}{c}{2} & \multicolumn{1}{c}{3} & \multicolumn{1}{c}{5} & \multicolumn{1}{c}{10}  \\
\hline
GCN & $59.4_{(6.7)}$        & $60.1_{(1.0)}$        & $60.4_{(4.6)}$        & $59.3_{(6.8)}$                 & $52.6_{(2.7)}$        & ${52.5}_{(2.5)}$        & $55.0_{(6.4)}$        & $54.5_{(3.9)}$                  \\
\hline
GSAE Acc   & $\mathbf{65.3}_{(5.5)}$        & $\mathbf{64.2}_{(7.9)}$        & $\mathbf{65.8}_{(6.2)}$        & $\mathbf{66.0}_{(5.3)}$                & $\mathbf{56.1}_{(4.6)}$        & $57.3_{(6.8)}$        & $\mathbf{59.1}_{(6.2)}$        & $\mathbf{61.3}_{(6.7)}$                \\
VGSAE Acc  & $64.6_{(4.2)}$        & $63.4_{(5.1)}$        & $65.6_{(6.0)}$        & $65.3_{(7.0)}$                 & $55.9_{(4.0)}$        & $\mathbf{58.7}_{(5.2)}$        & $57.8_{(6.5)}$        & $59.7_{(7.4)}$                  \\
\hline
GSAE Unc  & $\mathbf{64.6}_{(4.0)}$        & $63.5_{(7.2)}$        & $\mathbf{66.0}_{(5.3)}$        & $\mathbf{65.1}_{(6.1)}$              &  $\mathbf{58.0}_{(6.0)}$        & $57.4_{(7.1)}$        & $\mathbf{58.4}_{(6.1)}$        & $\mathbf{61.0}_{(7.0)}$               \\
VGSAE Unc  & $61.8_{(9.4)}$        & $\mathbf{63.7}_{(8.4)}$        & $65.2_{(5.2)}$        & $63.7_{(5.7)}$                 &  $55.2_{(3.9)}$        & $\mathbf{57.7}_{(6.6)}$        & $56.9_{(6.5)}$        & $59.4_{(5.8)}$                    \\
\hline
\end{tabular}}
\label{table:GAE}
\end{table}

\subsubsection{Impact of GNN Baseline}
We also conduct an ablation study to investigate the performance of our proposed GDM on additional GNN baselines. 
In particular, we applied GDM on 
the Graph Attention Network (GAT) \cite{velickovic2018graph} and Graph Isomorphism Network (GIN) \cite{xu2018how} baselines. The comparison results with multiple label rates, $\{2, 3, 5, 10\}$, on the IMDB-Binary and IMDB-Multi datasets are reported in Table \ref{table:GNN_Baseline}. The table clearly shows that GDM significantly improves the performance of both the GAT and GIN baselines across 
all label rates for both datasets. 
The performance gains are notable, exceeding 6\%, 5\% for GAT with label rates 5 and 3 for IMDB-Binary and IMDB-Multi, respectively. 
Similarly, GDM yields notable performance boost over GIN, exceeding 7\%, 5\% with label rates 3 and 5, respectively, for the IMDB-Binary dataset.

\begin{table}[t]\normalsize
\caption{Ablation study results on the impact of GNN baselines in terms of mean classification accuracy (standard deviation is within brackets).}
\resizebox{\textwidth}{!}{
\begin{tabular}{l|llll||llll}
\hline
  & \multicolumn{4}{c||}{IMDB-Binary}                                                         & \multicolumn{4}{c}{IMDB-Multi}                                                         \\

 & \multicolumn{1}{c}{2} & \multicolumn{1}{c}{3} & \multicolumn{1}{c}{5} & \multicolumn{1}{c||}{10} &  \multicolumn{1}{c}{2} & \multicolumn{1}{c}{3} & \multicolumn{1}{c}{5} & \multicolumn{1}{c}{10}  \\
         \hline
GIN      & $57.9_{(8.0)}$ & $54.5_{(6.1)}$ & $54.3_{(6.1)}$ & $56.7_{(9.7)}$ &  $32.6_{(4.8)}$ & $31.6_{(6.0)}$ & $32.8_{(4.6)}$ & $36.0_{(5.2)}$  \\
GDM Acc & $\textbf{58.8}_{(6.9)}$ & $57.0_{(5.5)}$ & $57.8_{(6.3)}$ & $\textbf{59.4}_{(6.3)}$ &  $35.6_{(4.2)}$ & $35.1_{(4.6)}$ & $36.9_{(3.9)}$ & $38.3_{(3.5)}$  \\
GDM Unc & $58.2_{(5.0)}$ & $\textbf{58.0}_{(6.5)}$ & $\textbf{60.5}_{(6.0)}$ & $57.7_{(6.2)}$ &  $\textbf{36.6}_{(4.2)}$ & $\textbf{37.2}_{(4.8)}$ & $\textbf{37.1}_{(3.9)}$ & $\textbf{39.5}_{(4.5)}$  \\
\hline
GAT      & $51.2_{(2.1)}$ & $50.6_{(1.2)}$ & $55.1_{(5.7)}$ & $54.4_{(5.5)}$ &$31.8_{(2.0)}$ & $34.0_{(1.5)}$ & $32.6_{(1.9)}$ & $33.6_{(2.4)}$  \\
GDM Acc & $\textbf{55.7}_{(4.3)}$ & $56.2_{(6.2)}$ & $55.0_{(6.5)}$ & $\textbf{60.0}_{(7.1)}$ & $35.6_{(2.9)}$ & $\textbf{36.4}_{(3.8)}$ & $\textbf{37.2}_{(4.3)}$ & $38.5_{(5.0)}$  \\
GDM Unc & $54.5_{(3.1)}$ & $\textbf{58.2}_{(6.7)}$ & $\textbf{56.0}_{(6.9)}$ & $59.2_{(5.0)}$ &$\textbf{35.8}_{(2.8)}$ & $34.4_{(3.6)}$ & $36.8_{(5.4)}$ & $\textbf{38.8}_{(4.3)}$ \\
\hline
\end{tabular}}
\label{table:GNN_Baseline}

\end{table}

\subsubsection{Impact of Graph Readout Method}

We conduct an ablation study to investigate the impact of the graph Readout function employed in our GNN model. Specifically, in addition to Global Mean Pooling, we consider the following two variants: (1) Add, where Global Add Pooling is used to obtain the graph-level embedding. (2) Max, where Global Max Pooling is used to obtain the graph-level embedding. The comparison results with different label rates on the Proteins and IMDB-Binary datasets are reported in Table \ref{table:Pool}. From the table, we can see that 
the Global Max Pooling and Global Add Pooling variants have performance drops compared to the Global Mean Pooling with almost all label rates for both 
the Proteins and IMDB-Binary datasets. Global Add Pooling suffers from obtaining un-normalized graph-level embeddings which causes generalization issues given that the graphs in each dataset have different sizes. Global Max Pooling only considers one feature per node corresponding to the feature with max value, causing the obtained graph-level embeddings to omit discriminative information present in the other features of the node-level embeddings. 

\begin{table}[t]\normalsize
\caption{Ablation study results on the impact of graph readout function in terms of mean classification accuracy (standard deviation is within brackets).}
\resizebox{\textwidth}{!}{
\begin{tabular}{l|llll||llll}
\hline
 & \multicolumn{4}{c||}{Proteins}                                                                                                                     & \multicolumn{4}{c}{IMDB-Binary}\\
 & \multicolumn{1}{c}{2} & \multicolumn{1}{c}{3} & \multicolumn{1}{c}{5} & \multicolumn{1}{c||}{10}    &\multicolumn{1}{c}{2} & \multicolumn{1}{c}{3} & \multicolumn{1}{c}{5} & \multicolumn{1}{c}{10} \\
\hline
 Acc Mean & $\mathbf{65.3}_{(5.5)}$        & $\mathbf{64.5}_{(4.8)}$        & $\mathbf{65.8}_{(5.2)}$        & $\mathbf{66.0}_{(5.3)}$                 & $\mathbf{56.1}_{(4.6)}$        & $\mathbf{57.3}_{(6.8)}$        & $\mathbf{59.1}_{(6.2)}$        & $\mathbf{61.3}_{(6.7)}$               \\
 Acc  Add  & $64.9_{(5.1)}$        & $63.2_{(4.9)}$        & $65.3_{(4.8)}$        & $63.3_{(4.3)}$                    & $53.0_{(2.8)}$        & $55.2_{(5.3)}$        & $56.9_{(6.1)}$        & $59.9_{(6.7)}$                   \\
 Acc  Max  & $63.8_{(4.7)}$        & $64.4_{(4.8)}$        & $63.2_{(7.0)}$        & $63.5_{(7.4)}$                      & $53.9_{(4.7)}$        & $56.0_{(7.8)}$        & $58.6_{(7.0)}$        & $60.6_{(6.3)}$                    \\
\hline
Unc Mean & $\mathbf{64.6}_{(4.0)}$        & $\mathbf{63.6}_{(6.0)}$        & $\mathbf{65.1}_{(5.3)}$        & ${66.0}_{(5.0)}$                    & $\mathbf{58.0}_{(6.0)}$        & $\mathbf{57.4}_{(7.1)}$        & $\mathbf{58.4}_{(6.1)}$        & $\mathbf{61.0}_{(7.0)}$                  \\
Unc Add  & $61.7_{(6.5)}$        & $\mathbf{63.6}_{(6.1)}$        & $63.0_{(7.0)}$        & $63.4_{(5.5)}$                   & $55.6_{(4.9)}$        & $54.8_{(3.6)}$        & $57.4_{(8.2)}$        & $60.5_{(6.7)}$                       \\
Unc Max  & $63.0_{(1.1)}$        & $63.4_{(6.6)}$        & $61.9_{(7.0)}$        & $\mathbf{66.8}_{(5.9)}$                      & $54.0_{(4.2)}$        & $54.4_{(7.6)}$        & $57.9_{(7.1)}$        & $60.4_{(7.1)}$                     \\
\hline
\end{tabular}}
\label{table:Pool}
\end{table}

\section{Conclusion}

In this paper, we proposed a novel Graph Dual Mixup (GDM) augmentation method 
for graph classification with limited labeled data. 
The proposed method employs a Graph Structural Auto-encoder to learn the structural embedding of the nodes, and then applies dual mixup on the structural node embeddings and the original node features of a pair of existing graphs in parallel to generate the structural and functional information of a new graph instance. 
The generated graph samples can augment the set of original graphs to alleviate overfitting and improve the generalizability of the GNN models. 
Additionally, we further propose two novel Balanced Graph Sampling methods to support GDM and enhance
the balanced difficulty and diversity of the generated graph samples. 
We conducted experiments on six graph benchmark datasets, the experimental results demonstrate that the proposed method improves
the generalization performance of the underlying GNNs when the labeled graphs are scarce and outperforms the state-of-the-art graph augmentation methods.

\bibliographystyle{splncs04}
 \bibliography{biblography}

\end{document}